\documentclass[journal]{IEEEtran}
\usepackage{amsmath,amsfonts}
\usepackage{url}
\usepackage{verbatim}
\usepackage{cite}
\usepackage{textcomp}
\usepackage{xcolor}
\usepackage[ruled,vlined,linesnumbered,commentsnumbered]{algorithm2e}
\usepackage{fancyvrb}
\usepackage{booktabs}
\usepackage{physics}
\usepackage{tikz}
\usepackage[hidelinks]{hyperref}
\usepackage{multirow}
\usepackage{pgfplots}
\pgfplotsset{compat=1.18} 
\usepackage{pdflscape}
\usepackage{flushend}

\newcommand{\mytriangle}[1]{\tikz{\filldraw[draw=#1,fill=#1] (0,0) --
(0.6em,0) -- (0.3em,0.6em);}}
\newcommand{\LGAFFS}{LGAFFS}
\newcommand{\PGAFFS}{PGAFFS}
\newcommand{\pred}{\hat{Y}}
\newcommand{\predl}{\hat{y}}

\begin{document}

\title{Fair Feature Selection: A Comparison of Multi-Objective Genetic Algorithms}

\author{\IEEEauthorblockN{James Brookhouse\IEEEauthorrefmark{1},
Alex Freitas\IEEEauthorrefmark{2} \\ 
\IEEEauthorblockA{School of Computing,
University of Kent\\
Canterbury, United Kingdom\\
Email: \IEEEauthorrefmark{1}james@brookhou.se,
\IEEEauthorrefmark{2}A.A.Freitas@kent.ac.uk
}}}

\maketitle

\begin{abstract}
Machine learning classifiers are widely used to make decisions with a major impact on people's lives (e.g. accepting or denying a loan, hiring decisions, etc). In such applications, the learned classifiers need to be both accurate and fair with respect to different groups of people, with different values of variables such as sex and race. This paper focuses on fair feature selection for classification, i.e. methods that select a feature subset aimed at maximising both the accuracy and the fairness of the predictions made by a classifier. More specifically, we compare two recently proposed Genetic Algorithms (GAs) for fair feature selection that are based on two different multi-objective optimisation approaches: (a) a Pareto dominance-based GA; and (b) a lexicographic optimisation-based GA, where maximising accuracy has higher priority than maximising fairness. Both GAs use the same measures of accuracy and fairness, allowing for a controlled comparison. As far as we know, this is the first comparison between the Pareto and lexicographic approaches for fair classification.
The results show that, overall, the lexicographic GA outperformed the Pareto GA with respect to accuracy without degradation of the fairness of the learned classifiers. This is an important result because at present nearly all GAs for fair classification are based on the Pareto approach, so these results suggest a promising new direction for research in this area.
\end{abstract}


\section{Introduction}

Historically, when developing algorithms for the classification task of machine learning, the main focus has been on predictive performance, with a moderate emphasis on the simplicity of classification models (classifiers) through metrics such as the size of models produced -- as a proxy to the more complex concept of comprehensibility or interpretability \cite{Freitas2014}, \cite{Burkart2021}. Recently, the fairness of machine learning algorithms has come under increasing scrutiny \cite{mehrabi2019survey}, \cite{Binns2018}, \cite{vanGiffen2022}. For example ProPublica investigated the COMPASS recidivism algorithm that was used to categorise the risk of criminals going on to re-offend, finding that the algorithm was much more likely to classify black defendants as higher risk \cite{propublica2016}.

Fairness is a complex concept and no single measure can capture all the nuances of what it means to be fair. Because of this, the field has created a large number of measures that each capture different notions of what it means to be fair \cite{verma2018fairness}. In fact some of these measures contradict each other and optimising for one measure can be at the detriment of another \cite{kleinberg2016inherent},  \cite{chouldechova2017fair}. 

In this paper, we tackle the feature selection task, which consists of selecting a subset of features for a classifier in a pre-processing step, i.e., before learning the final classifier.
The objective is to select a subset of features that optimises both the predictive accuracy and the fairness of the models (classifiers) learned by the subsequent classification algorithm.

The main contribution of the paper is to compare two types of recently proposed Genetic Algorithms (GAs) for multi-objective feature selection (optimising accuracy and fairness): one based on the Pareto dominance concept \cite{Rehman2022} and another based on the concept of lexicographic optimisation \cite{brookhouse2022}. This type of comparison is new and important because, although the Pareto dominance approach is much more often used in the multi-objective GA literature than the lexicographic approach, the latter produced better results overall in our experiments, which suggests that the lexicographic approach deserves more attention in the literature.

The rest of this paper is organised as follows. First, Section \ref{sec:background} presents the background on fairness and how it applies to machine learning, and some background on multi-objective optimisation. Section \ref{sec:algorithms} discusses the two Genetic Algorithms (GAs) for fair feature selection compared in this work. Section \ref{sec:setup} describes the experimental setup to perform a controlled comparison between these two GAs. Section \ref{sec:results} reports the experimental results, before an analysis from both the lexicographic and Pareto perspectives in Section \ref{sec:Discussion}. Section \ref{sec:Conclusions} presents the conclusions.

\section{Background}
\label{sec:background}

In this section, we will introduce fairness in machine learning and different approaches to measure fairness. We will then discuss the two multi-objective approaches evaluated in this paper: the Pareto dominance and the lexicographic approaches.

\subsection{Fairness in Machine Learning}

What does it mean to be fair in machine learning applications? If we consider loan applications, for example, we could consider a fair model to be one that grants the same number of loans to all groups of individuals (e.g. male and female groups), or it could be fair if the probability of acceptance is the same across different groups. However, these two ideas look at the larger picture and do not measure the effect at an individual level, where a model would be fair if its predictions were consistent between similar individuals.  

Before discussing further what it means to be fair, we first have to introduce some terminology. In the classification task, datasets contain a number of features (predictive variables) that are used to make predictions about a target (class) variable. When considering fairness, some of the predictive features are also known as sensitive (or protected) features. For example, gender, race, and age could be considered sensitive features. The values of sensitive features can be used to split individuals (records in a dataset) into two groups: protected and unprotected individuals. The protected group contains the individuals that are considered to be subject to unfair bias and are more likely to obtain a negative outcome (class label) than the unprotected group.

Many fairness measures have been proposed that represent some notion of fairness in a model learned from data \cite{mehrabi2019survey,verma2018fairness}. These measures can be grouped into two main categories: individual-level and group-level measures.

One group-level fairness measure is demographic parity (or discrimination score) \cite{mehrabi2019survey}, \cite{calders2010three}. This measures the difference between the predicted positive-class probabilities of the protected and unprotected groups. Other group-level metrics look at specific elements of a classifier's confusion matrix, such as the number of false positives and false negatives in both the protected and unprotected groups \cite{chouldechova2017fair}. One limitation of group-level fairness measures in general is that they do not consider the individual. Hence, two similar individuals either within a group or across groups could receive a different outcome, which would be unfair at the individual level. This fine-grained notion of fairness is not captured by group-level fairness measures.

To overcome this limitation, individual-level fairness measures have been proposed. One such measure is consistency. Consistency uses a k-nearest neighbour approach to measure how many of an individual's nearest neighbours have the same outcome (predicted class) as it. If an individual has the same predicted class as its nearest neighbours, those predictions are maximally consistent. This test is performed for each individual (i.e., each instance in the training set) and the average taken \cite{zemel2013learning}. One limitation of consistency is that as the number of features grows, the distances between instances (individuals) also tends to grow and the notion of ``nearest neighbours'' becomes increasingly diluted.

Hence, there is no single measure of fairness that can capture all nuances of what it means to be fair. For this reason, the algorithms presented in Section \ref{sec:algorithms} optimise multiple measures of fairness, as well as optimising a measure of predictive accuracy.

\subsection{Multi-Objective Optimisation}

There are two main multi-objective approaches in machine learning: the Pareto dominance approach (by far the most popular in the area of evolutionary algorithms) and the lexicographic approach. In this section, we will briefly review both these approaches first, before discussing their pros and cons.

\subsubsection{Pareto Multi-Objective Optimisation}

This approach aims at finding a set of non-dominated individuals, based on the concept of Pareto dominance. An individual (candidate solution) $I_i$  dominates another individual $I_j$ if $I_i$ is superior to $I_j$ in at least one objective and $I_i$ is no worse than $I_j$ in all other objectives. An individual $I_i$ is deemed non-dominated if there is no other individual in the population that dominates $I_i$. At each iteration of the GA, every candidate solution (individual) generated is tested to check if it is a non-dominated solution. 

\begin{figure}[b]
    \centering
\begin{tikzpicture}
\begin{axis}[
    enlargelimits=false,
    xmin=0,
    xmax=10,
    ymin=0,
    ymax=10,
]
\addplot[
    color=blue,
    mark=triangle
    ]
table[]
{non_dom.dat};

\addplot[
    color=blue,
    only marks,
    mark=triangle*
    ]
table[]
{non_dom.dat};

\addplot[
    color=red,
    only marks,
    ]
table[]
{dom.dat};

\end{axis}
\end{tikzpicture}
    \caption{Example Pareto front for two objectives to be maximised. The non-dominated solutions are shown in blue triangles and the dominated solutions in red circles.}
    \label{fig:pareto_front}
\end{figure}
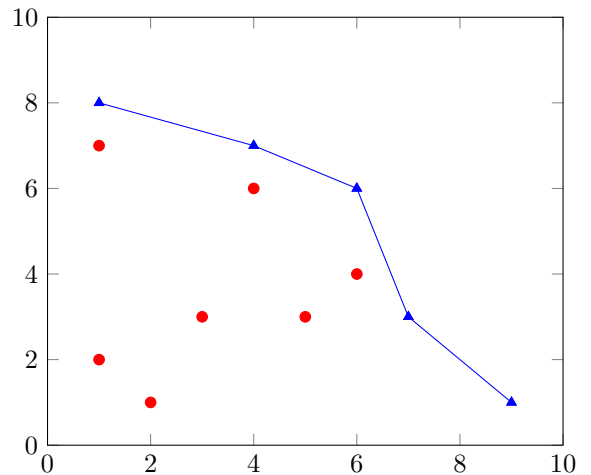

Figure \ref{fig:pareto_front} shows an example set of candidate solutions for a two-objective problem -- each objective is represented as an axis on the graph. Assuming, without loss of generality, that both objectives are to be maximised, there is a set of dominated solutions (red circles) and a set of non-dominated solutions (blue triangles) which are forming the Pareto front. A very popular family of GAs using Pareto dominance for multi-objective optimisation is the NSGA (Non-dominated Sorting Genetic Algorithm) series of algorithms \cite{srinivas1994nsga,deb2002nsga2}. 

Several GAs for fair machine learning have been recently proposed based on the Pareto dominance concept, typically using NSGA-II as the multi-objective optimisation algorithm. This includes the GA for optimising the hyper-parameters of a fair classifier proposed in \cite{valdivia2021}, the GA for finding fair counterfactuals proposed in \cite{Dandl2022}, a version of the Genetic Programming method for fair feature construction proposed in \cite{laCava2020}, and the GA for fair feature selection proposed in \cite{Rehman2022}. 

The GA in \cite{Rehman2022} is particularly relevant for this current work, since it focuses on our target problem of fair feature selection. Hence, this GA, based on Pareto dominance, will be somewhat modified and compared against a lexicographic optimisation-based GA for feature selection (see next Subsection) in controlled experiments reported later in this paper.

\subsubsection{Lexicographic Multi-Objective Optimisation}

This approach requires that an order of importance (priority) for the objectives be defined by the user. It considers each objective in turn, starting with the most important objective in the ordered list of objectives. Let $I_i$ and $I_j$ be two individuals being compared, and let $V_i$ and $V_j$ be their values for the current objective. If the absolute value of the difference between $V_i$ and $V_j$ is greater than a user-defined very small threshold epsilon -- i.e., if $|V_1 - V_2| > \epsilon$ -- then the individual with the best value for that objective is selected, and there is no need to consider the values of other objectives for $I_i$ and $I_j$. However, if $|V_i - V_j|$ is within $\epsilon$, the difference between those two objective values is deemed irrelevant (negligible), i.e. $I_i$ and $I_j$ are equivalent with respect to the current objective; therefore, the individuals $I_i$ and $I_j$ are further compared with respect to the next objective in the ordered list, and so on. If $I_i$ and $I_j$ are equivalent with respect to all objectives, then the individual with the best value of the most important objective is selected.

A different $\epsilon$ could be assigned to each objective. This would be particularly useful, e.g., if the ranges of valid values differs between objectives.

One of the earliest machine learning algorithms using the lexicographic approach was the classical rule induction algorithm AQ18 \cite{kaufman1999learning}. AQ18 optimises two objectives, with each objective having its own $\epsilon$ (called tolerance in that work). The first, highest priority  objective is an accuracy measure, while the second, lower priority objective is a rule simplicity measure.

Another machine learning algorithm using the lexicographic approach is LEGAL-tree \cite{basgalupp2009legal}, which learns decision trees from data using a lexicographic GA. In essence, each individual represents a candidate decision tree, and selection is performed by lexicographic tournament selection. LEGAL-tree uses two objectives: the first (higher priority) one is predictive accuracy, and the secondary (lower priority) one is the size of the decision tree, as a measure of model simplicity. 

In the specific context of fairness in classification, in addition to the lexicographic-optimisation GA for fair feature selection proposed in \cite{brookhouse2022}, which is evaluated in this paper, we are aware of only one related work that is broadly based on the lexicographic approach: the Genetic Programming (GP) algorithm for feature construction for fair classification proposed in \cite{laCava2020}. That GP uses lexicase selection, which follows the broad lexicographic principle of optimising performance across an ordered list of sub-groups of instances. However, in that GP the lexicographic orderings of sub-groups of instances are $randomly$ generated, instead of reflecting a user-specified order of priority for the objectives, as defined for the lexicographic approach in this paper.

\subsubsection{Pros and Cons of the Pareto and Lexicographic approaches}
\label{sec:pros_cons_pareto_lexic}

The Pareto and lexicographic approaches have to some extent complementary pros and cons, as follows.

First, the Pareto approach has the advantage of not requiring any parameters related to how to cope with the trade-off between the different objectives. Recall that the lexicographic approach requires a priority order of objectives, and it introduces a new epsilon parameter(s) whose value(s) have to be specified by the user to define whether a difference in objective values (between two individuals) is relevant. The Pareto approach does not require these epsilon parameters and also does not need any ranking of objectives. For some problems, a natural ranking may not be known before the algorithm is ran, or a sensible ranking may not exist. 

On the other hand, in some applications there is a natural order of importance (priority) for the objectives to be optimised. For example, in the classification task of machine learning, there is a general consensus that optimising predictive accuracy is more important than optimising other objectives like the simplicity of the learned model or the running time of the classification algorithm. 
In applications where the user wants to specify a priority order for objectives, it is important to give this kind of background knowledge to the optimisation algorithm; and this is easily achieved with the lexicographic approach. Note that it is usually easier and more natural for users to specify an ordering of objectives than a specific numeric value for the weight of each objective.

By contrast, the Pareto approach does not allow users to specify a priority order of objectives for the GA. In this approach, any priority order would need to be considered later, when analysing the many non-dominated solutions returned by the GA. 
Note that ignoring the user's priority order among objectives leads to some computational inefficiency in the search performed by a Pareto-based GA, due to potentially selecting and preserving in the population, for many generations, solutions that would not be selected by the user in the end, due to a clearly bad trade-off between the objectives. For example, if the user specifies that optimising a classifier's predictive accuracy has priority over optimising its size (simplicity), an individual with the smallest size and very low accuracy could be selected and remain in the Pareto front for many generations, despite being clearly a bad solution. The lexicographic approach would never select that bad solution due to its very low accuracy (as the highest-priority objective).

The Pareto and lexicographic approaches also have different pros and cons in terms of the number of solutions returned to the user, as follows. The Pareto approach returns a front of non-dominated solutions, with no solution on this front being considered better or the best. So, users are still left with the problem of how to choose the best solution out of many (often very many) non-dominated solutions. This gives users the opportunity to consider the trade-offs between different objectives among many non-dominated solutions and choose the best solution based on their subjective preferences; however, it requires users to spend time analysing the returned Pareto front. By contrast, the lexicographic approach returns a single best individual (solution) to the user. This saves time for the user, but it does not allow users to manually choose the best solution based on their subjective preferences.

\section{\LGAFFS{} and \PGAFFS{}}
\label{sec:algorithms}

This paper compares two recently proposed multi-objective Genetic Algorithms (GAs) for fair feature selection in the classification task: our Lexicographic GA for Fair Feature Selection (\LGAFFS{}) \cite{brookhouse2022} and a Pareto dominance-based feature selection GA \cite{Rehman2022} that uses the well known NSGA-II algorithm as its base \cite{deb2002nsga2}. Note that, to make this comparison as controlled and as ``fair'' as possible, we had to modify some characteristics of the Pareto GA in \cite{Rehman2022}, as will be explained later.
First, however, we will define the fairness and predictive accuracy measures being optimised by both GAs compared in this work in Sections \ref{sec:metrics} and \ref{sec:accuracy_metric}, respectively. Then, we will describe the lexicographic GA (\LGAFFS{}) in Section \ref{sec:LGAFFS} and the Pareto GA (\PGAFFS{}) in Section \ref{sec:pareto}.

\subsection{Measures of Fairness}
\label{sec:metrics}

No single fairness measure captures all nuances of a fair model, so both the lexicographic and Pareto approaches optimise four fairness measures to get more robust fairness results. The fairness measures used each capture a different concept of what it means to be fair, including both individual and group level fairness as well as covering both positive and negative class prediction imbalances.

To define these measures (which were also used in \cite{brookhouse2022}) we use the following terminology and notation:

\begin{itemize}
	\item $S$: Protected/sensitive feature: taking value 0 (unprotected group) or 1 (protected group)
	\item $\pred$ : the predicted class; $Y$: the actual class; taking class label 1 (positive class) or 0 (negative class)
	\item $TP$, $FP$, $TN$, $FN$: Number of True Positives, False Positives, True Negatives and False Negatives, respectively
\end{itemize}

\noindent The first measure is demographic parity (DP) (or discrimination score) \cite{mehrabi2019survey}, \cite{calders2010three}, which is defined as:

\begin{equation}
\label{eq:discrimination_score}
DP = 1 - \abs{P(\pred = 1 \rvert S = 0) - P(\pred = 1 \rvert S = 1)}
\end{equation}

\noindent DP is a group-level fairness measure that takes the optimal value of 1 if both protected and unprotected groups have an equal probability of being assigned to the positive class by the classifier. Note that on unbalanced-class datasets, where there is a large difference between the relative frequencies of the positive class for the protected and unprotected groups, maximising DP tends to reduce predictive accuracy.

The second measure used is consistency\cite{zemel2013learning}, defined as:

\begin{equation}
\label{eq:consistency}
C =  1 - \frac{1}{Nk}\sum_i{\sum_{j \in kNN(x_i)}{\abs{\predl_i - \predl_j}}}
\end{equation}

\noindent where $N$ is the number of training instances. Consistency is an individual-level similarity metric that compares the class predicted by a classifier to each instance in the dataset to the class predicted by the classifier to that instance's $k$ nearest instances (neighbours) in the dataset. If all these neighbours have the same predicted class as the current instance, then the class prediction for that instance is considered consistent. The measure computes the average degree of consistency over all training instances. A fully consistent model has a consistency of 1 and a fully inconsistent model has a value of 0. 

Thirdly, the False Positive Error Rate Balance Score (FPERBS) \cite{chouldechova2017fair, corbett2018measure} is:
\begin{equation}
\label{eq:FPERBS}
FPERBS = 1- \abs{\frac{FP_{S=0}}{FP_{S=0}+TN_{S=0}} - \frac{FP_{S=1}}{FP_{S=1}+TN_{S=1}}}
\end{equation}

\noindent FPERBS measures the difference in the probability that a truly negative instance is incorrectly predicted as positive between protected and unprotected groups. 

Fourthly, the False Negative Error Rate Balance Score (FNERBS) \cite{chouldechova2017fair,hardt2016equality,kusner2017counterfactual} measures the difference in the probability that a truly positive instance is incorrectly predicted as negative between protected and unprotected groups:

\begin{equation}
\label{eq:FNERBS}
FNERBS = 1 - \abs{\frac{FN_{S=0}}{FN_{S=0}+TP_{S=0}} - \frac{FN_{S=1}}{FN_{S=1}+TP_{S=1}}}
\end{equation}

\noindent A score of 1 indicates an optimally fair result for both FPERBS and FNERBS. Both FPERBS and FNERBS are group-level measures of fairness.

\subsection{Measure of Predictive Accuracy}
\label{sec:accuracy_metric}

As the accuracy measure to be optimised, both the lexicographic and the Pareto GAs evaluated in our experiments use the geometric mean of Sensitivity and Specificity (Equations \ref{eq:accuracy_measure} and \ref{eq:combined_accuracy}). 
This measure was chosen because it incentivises the correct classification of both positive-class and negative-class instances, to counteract pressure from the fairness measures to produce maximally fair models that trivially predict the same class for all instances.

\begin{equation}
\label{eq:accuracy_measure}
\begin{split}
Sensitivity &= \frac{TP}{TP + FN}, \\ \quad \quad Specificity &= \frac{TN}{TN + FP}
\end{split}
\end{equation}

\noindent
Sensitivity and Specificity are combined to give the geometric mean (GM) of these two measures:
\begin{equation}
\label{eq:combined_accuracy}
    GM_{Sen \times Spec} = \sqrt{Sensitivity \cdot Specificity}
\end{equation}

\subsection{The Lexicographic GA for Fair Feature Selection (LGAFFS)}
\label{sec:LGAFFS}

\LGAFFS{} \cite{brookhouse2022} is a recently proposed lexicographic-optimisation GA that uses the wrapper approach for fair feature selection.

\LGAFFS{} uses a standard individual representation for feature selection, where an individual consists of $F$ bits, where $F$ is the number of features in the original dataset, and each bit indicates whether or not its corresponding feature is selected.

\LGAFFS{} uses a lexicographic tournament selection, where the higher-priority objective is to maximise the $GM_{Sen \times Spec}$ measure (equation \ref{eq:combined_accuracy}), as a measure of predictive accuracy; and the lower-priority objective is to maximise the four fairness measures defined in Section \ref{sec:metrics}. Since none of those four fairness measures can be considered more important than the others, the GA's tournament selection procedure aggregates the values of the four fairness measures into a single objective, as follows. 

The GA generates all possible 24 (4!) possible permutations of the four fairness measures, where each permutation specifies a priority order for the four fairness measures to be optimised. Then, each of those orders (permutations) is used to perform a lexicographic comparison between the fairness measures' values of the two individuals in the tournament, comparing the individuals with respect to one fairness measure at a time until a substantial difference (greater than a threshold $\epsilon$) is found, when the best individual is deemed the winner for that permutation. Hence, if the tournament is not decided based on the $GM_{Sen \times Spec}$ measure, the tournament winner is the individual that won in the largest number of those permutations of fairness measures -- see \cite{brookhouse2022} for more details.

The high-level pseudo-code for \LGAFFS{} is shown in Algorithm \ref{algo:LGAFFS}. Population initialisation is performed by a procedure that promotes an increased diversity in the population, as the probability of randomly turning on (selecting) a feature varies between individuals in the population, so that different individuals tend to have quite different numbers of selected features. 

The fitness values of each individual -- i.e. the values of $GM_{Sen \times Spec}$ and the four fairness measures -- are computed by a well-known internal cross-validation applied to the training set only (i.e. not using the test set). This internal cross-validation is used in lines 6 and 7 of Algorithm \ref{algo:LGAFFS}.

Line 10 of Algorithm \ref{algo:LGAFFS} implements a process of ``lexicographic elitism'', i.e., it selects the best individual of the population, based on the previously discussed concept of lexicographic optimisation, and the selected individual is passed unaltered to the next generation. After this elitism, the previously discussed lexicographic tournament selection is applied to select parents for reproduction, and new children are created by using standard uniform crossover and bit-flip mutation operators (lines 11-15 of Algorithm \ref{algo:LGAFFS}). 

\LGAFFS{} has previously been compared against a sequential forward (greedy) search method for feature selection and against the baseline of no feature selection in a pre-processing phase (i.e. using all features to train the final classifier).
In both cases \LGAFFS{} showed a similar level of performance with respect to the accuracy measure, and it statistically improved several fairness measures in those experiments. In order to further evaluate \LGAFFS{} we will be comparing its performance against the performance of a Pareto dominance-based GA, as described in the next Subsection.

\begin{algorithm}
\normalsize
\SetKwProg{Fn}{Function}{:}{}
\SetKw{In}{in}
\SetKw{Or}{or}
\DontPrintSemicolon
\Fn{LGAFFS()}{
    population = initialise\_ramped\_population()\;
    folds = create\_folds()\;
    \For{$i$ \KwTo $max\_iterations$}{
        \ForEach{$individual$ \In $population$}{
            run\_classifier\_with\_cross\_validation(folds)\;
            calculate\_fitness\_measures()\;
        }
        best\_individual = population\_lexicographic\_top()\;
        new\_population = \{\}\;
        new\_population.append(best\_individual)\;       
        \While{size(new\_population) $<$ pop\_size}{
            i1, i2 = lexicographic\_tournament\_selection(population)\;
            i1', i2' = uniform\_crossover(i1, i2)\;
            bit-flip\_mutation(i1', i2')\;
            new\_population.append(i1', i2')\;
        }
        population = new\_population\;
    }
    \Return{best\_individual}
}
\caption{High-level pseudo-code for \LGAFFS{}. \label{algo:LGAFFS}}
\end{algorithm}

\subsection{The Pareto GA for Fair Feature Selection (PGAFFS)}
\label{sec:pareto}

The Pareto Genetic Algorithm for Fair Feature Selection (\PGAFFS{}) used in our experiments is a somewhat modified version of the GA proposed in \cite{Rehman2022}, which is in turn based on the very popular NSGA-II algorithm \cite{deb2002nsga2}. 

We preserved as much as possible the characteristics of the GA in \cite{Rehman2022}, including its individual representation, uniform crossover and mutation operators (which are all broadly the same as used in \LGAFFS{}) and the Pareto dominance-based multi-objective optimisation procedure (using NSGA-II). 

However, in order to make the comparison between \PGAFFS{} and \LGAFFS{} as controlled and as ``fair'' as possible, we modified the original GA in \cite{Rehman2022} in two ways: (a) we replaced its original accuracy and fairness measures (fitness functions) by the same accuracy and fairness measures used by \LGAFFS{}; and (b) we extended the GA in \cite{Rehman2022} to use the same population initialisation procedure used by \LGAFFS{}, which promotes higher diversity of individuals \cite{brookhouse2022}.

Hence, \PGAFFS{} aims to optimise two objectives (measures): the accuracy-based geometric mean of Sensitivity and Specificity (the same as optimised by \LGAFFS{}), and an aggregated fairness measure, which combines the same four fairness measures used by \LGAFFS{} -- with a difference in how the aggregation is performed by the two GAs, as follows. In \LGAFFS{}, fairness aggregation is performed as a lexicographic tournament when comparing two individuals (described in the previous Subsection). This approach is incompatible with the Pareto approach and NSGA-II, as individuals are not ranked in the same way. Instead, \PGAFFS{} simply aggregates the four fairness metrics into a single objective by computing their arithmetic mean (note that all four fairness measures produce values between 0 and 1 and they are all to be maximised). 

It is important to emphasise that \PGAFFS{} uses the same individual representation, the same standard genetic operators of uniform crossover and bit-flip mutation, and the same population-initialisation procedure used by \LGAFFS{} (as described in the previous Section).

Hence, \PGAFFS{} differs from \LGAFFS{} mainly regarding their multi-objective approaches (Pareto vs. lexicographic approaches). Thus, the two GAs use different selection procedures and, although they optimise the same predictive accuracy measure and the same four fairness measures, the way those fairness measures are aggregated into a single objective value differs between the two GAs, as described earlier.

\section{Experimental Setup}
\label{sec:setup}
In this section we describe the datasets used in all experiments along with the settings of the GAs' parameters used in the experiments.

\begin{table}[tb]
	\centering
	\caption{Datasets used in all experiments, detailing the number of instances, number of features and the sensitive features for each dataset. \label{tab:datasets}}
	\begin{tabular}{@{}llll@{}}
		\toprule
		Data set                 & Instances                     & Features   & Sensitive             \\
		                         &                               &            & Features              \\ 
		\midrule
		Adult Income (US Census) & 48842                         & 14         & Race, Gender,         \\
		                         &                               &            & Age                   \\
		German Credit            & 1000                          & 20         & Age, Gender           \\
		                         &                               &            &                       \\
		Credit Card Default      & 30000                         & 24         & Gender                \\
		                         &                               &            &                       \\
		Communities              & 1994                          & 128        & Race                  \\
		                         &                               &            &                       \\
		Student Performance (Portuguese)    & 650                & 30         & Age, Gender,          \\
		                         &                               &            & Relationship          \\
		Student Performance (Maths)         & 396                & 30         & Age, Gender,          \\
				                 &                               &            & Relationship          \\
		Recidivism               & 6167                          & 52         & Race, Gender          \\
		\bottomrule
	\end{tabular}
\end{table}

\subsection{Datasets}

The experiments used seven real-world datasets. Six of these datasets were obtained from the well-known UCI Machine Learning Repository \cite{Dua2019}, while the ProPublica Recidivism dataset was released by ProPublica \cite{propublica2016} during an investigation into biases in the classification of offenders. Table \ref{tab:datasets} lists these datasets along with their number of instances and features, and importantly the sensitive features that have been identified. All of these datasets have been used in prior work that has investigated fairness in machine learning. Note that most datasets are associated with multiple sensitive features. In these cases, each pair of a dataset and a sensitive feature is a unique classification problem, an independent experiment. Hence, the experiments used in total 21 classification problems, as shown in the tables of results.

\subsection{Experimental GA Parameters}

In order to make the comparison between \LGAFFS{} and \PGAFFS{} as controlled and as fair as possible, both GAs were set with the same values for the parameters that they have in common. These shared GA parameters were set as follows:

\begin{itemize}
   \item General GA parameters: population\_size: 101, MAX\_P: 0.5, MIN\_P: 0.1 (MAX\_P and MIN\_P are population-initialisation parameters, see \cite{brookhouse2022}),  number of folds for internal cross-validation: 3, max\_iterations: 50, tournament\_size: 2, crossover\_probability: 0.9, mutation\_probability: 0.05
\end{itemize}

In addition, \LGAFFS{} has its own specific parameters, which were set as follows:

\begin{itemize}
    \item \LGAFFS{}'s lexicographic-optimisation-specific parameters: accuracy\_$\epsilon$ and fairness\_$\epsilon$: 0.01, fair\_rank\_$\epsilon$ and fair\_test\_$\epsilon$: 1 (again, see \cite{brookhouse2022} for details)
    
\end{itemize}

Finally, in all experiments the features selected by the GAs are used as input to a random forest classification algorithm, which is one of the most powerful and popular types of classification algorithms \cite{breiman2001random},\cite{Fernandez-Delgado2014}, \cite{Zhang2017}.

\begin{table*}[t]
\caption{Comparison of the individual produced by \LGAFFS{} and an individual from the \PGAFFS{}'s Pareto front selected by a lexicographic filter (where accuracy has priority over fairness). The Wilcoxon signed-ranked test was applied to each measure and the results are shown at the bottom of each column. Statistically significant results are indicated with a red triangle.\label{tab:comparison_lex}}
\centering
\footnotesize
\begin{tabular}{@{}llllllllllll@{}}
\toprule
                       & Sensitive & \multicolumn{2}{l}{$GM_{Sen \times Spec}$} & \multicolumn{2}{l}{Demographic Parity} & \multicolumn{2}{l}{Consistency} & \multicolumn{2}{l}{FPERBS} & \multicolumn{2}{l}{FNERBS} \\ \cmidrule(l){3-4} \cmidrule(l){5-6} \cmidrule(l){7-8} \cmidrule(l){9-10} \cmidrule(l){11-12}    
Data Set               & Feature & \PGAFFS{}          & \LGAFFS{}          & \PGAFFS{}                & \LGAFFS{}                & \PGAFFS{}           & \LGAFFS{}            & \PGAFFS{}         & \LGAFFS{}         & \PGAFFS{}         & \LGAFFS{}         \\ \midrule
Adult                  & age       & \textbf{0.7021}        & 0.6475       & 0.7530              & \textbf{0.8485}             & 0.7537         & \textbf{0.8656}         & 0.8796       & \textbf{0.9522}      & 0.5041       & \textbf{0.9189}      \\
Adult                  & sex       & 0.6979        & \textbf{0.7409}       & 0.8082              & \textbf{0.9356}             & 0.7482         & \textbf{0.8201}         & 0.8961       & \textbf{0.9862}      & 0.8641       & \textbf{0.9902}      \\
Adult                  & race      & 0.6566        & \textbf{0.7420}       & \textbf{0.9467}              & 0.8498             & 0.7677         & \textbf{0.8163}         & \textbf{0.9799}       & 0.9490      & \textbf{0.9649}       & 0.9416      \\
German Credit          & age       & 0.5855        & \textbf{0.5901}       & 0.8565              & \textbf{0.9361}             & 0.7220         & \textbf{0.7642}         & \textbf{0.9365}       & 0.8633      & 0.8707       & \textbf{0.8911}      \\
German Credit          & gender    & 0.4397        & \textbf{0.6036}       & 0.8546              & \textbf{0.9399}             & \textbf{0.7720}         & 0.7510         & \textbf{0.9000}       & 0.8993      & 0.8367       & \textbf{0.9230}      \\
Student Maths          & age       & 0.8439        & \textbf{0.9208}       & 0.6000              & \textbf{0.7964}             & 0.8200         & \textbf{0.8444}         & 0.5714       & \textbf{0.9022}      & \textbf{0.9130}       & 0.8951      \\
Student Maths          & dalc      & 0.4529        & \textbf{0.8951}       & \textbf{0.8462}              & 0.7563             & 0.8350         & \textbf{0.8377}         & 0.2308       & \textbf{0.8051}      & \textbf{0.8846}       & 0.8157      \\
Student Maths          & walc      & \textbf{0.9245}        & 0.9052       & \textbf{0.9373}              & 0.7039             & 0.7900         & \textbf{0.8361}         & \textbf{0.9000}       & 0.8760      & \textbf{0.9588}       & 0.9222      \\
Student Maths          & famrel    & 0.8867        & \textbf{0.8984}       & 0.3684              & \textbf{0.9076}             & \textbf{0.8750}         & 0.8468         & 0.9091       & \textbf{0.9151}      & 0.8519       & \textbf{0.9106}      \\
Student Maths          & romantic  & 0.8851        & \textbf{0.9027}       & \textbf{0.8407}              & 0.8397             & 0.7750         & \textbf{0.8387}         & \textbf{0.9762}       & 0.7646      & 0.9000       & \textbf{0.9012}      \\
Student Maths          & sex       & 0.8490        & \textbf{0.9000}       & 0.8120              & \textbf{0.8364}             & 0.8050         & \textbf{0.8376}         & \textbf{0.8611}       & 0.8206      & \textbf{0.9235}       & 0.9196      \\
Student Portuguese     & age       & \textbf{0.8367}        & 0.8196       & \textbf{0.9481}              & 0.8638             & \textbf{0.9138}         & 0.9106         & \textbf{0.9167}       & 0.7038      & \textbf{1.0000}       & 0.9578      \\
Student Portuguese     & dalc      & 0.7675        & \textbf{0.7867}       & \textbf{0.9606}              & 0.8470             & \textbf{0.9138}         & 0.9128         & 0.6000       & \textbf{0.6230}      & 0.8571       & \textbf{0.9088}      \\
Student Portuguese     & walc      & 0.7746        & \textbf{0.8031}       & \textbf{0.9197}              & 0.8019             & \textbf{0.9262}         & 0.9097         & 0.2000       & \textbf{0.6450}      & \textbf{1.0000}       & 0.9205      \\
Student Portuguese     & famrel    & 0.6876        & \textbf{0.7846}       & 0.8648              & \textbf{0.9370}             & 0.8923         & \textbf{0.9211}         & \textbf{1.0000}       & 0.7608      & 0.9434       & \textbf{0.9686}      \\
Student Portuguese     & romantic  & 0.7532        & \textbf{0.8110}       & \textbf{0.9923}              & 0.9200             & 0.8862         & \textbf{0.9134}         & 0.4286       & \textbf{0.7646}      & \textbf{0.9888}       & 0.9708      \\
Student Portuguese     & sex       & 0.7006        & \textbf{0.8035}       & \textbf{0.9740}              & 0.9271             & \textbf{0.9292}         & 0.9186         & 0.5833       & \textbf{0.7214}      & 0.9286       & \textbf{0.9679}      \\
Communities  & race      & 0.8246        & \textbf{0.8303}       & 0.5703              & \textbf{0.6623}             & \textbf{0.6230}         & 0.6056         & 0.7752       & \textbf{0.9182}      & 0.7953       & \textbf{0.8313}      \\
Default of Credit      & sex       & 0.4205        & \textbf{0.5896}       & \textbf{0.9823}              & 0.9755             & \textbf{0.8734}         & 0.8354         & \textbf{0.9764}       & 0.9739      & \textbf{0.9926}       & 0.9852      \\
Recidivisim & race      & 0.6850        & \textbf{0.7306}       & 0.8028              & \textbf{0.8324}             & \textbf{0.6861}         & 0.6849         & 0.8895       & \textbf{0.9022}      & 0.8389       & \textbf{0.9105}      \\
Recidivisim & sex       & 0.5984        & \textbf{0.7113}       & 0.9359              & \textbf{0.9408}             & \textbf{0.6947}         & 0.6756         & \textbf{0.9311}       & 0.9254      & \textbf{0.9779}       & 0.9451      \\ \midrule
\multicolumn{2}{c}{Wins}                      & \multicolumn{1}{c}{3}             & \multicolumn{1}{c}{18}           & \multicolumn{1}{c}{10}                  & \multicolumn{1}{c}{11}                 & \multicolumn{1}{c}{10}             & \multicolumn{1}{c}{11}             & \multicolumn{1}{c}{10}            & \multicolumn{1}{c}{11}          & \multicolumn{1}{c}{10}            & \multicolumn{1}{c}{11}          \\ \cmidrule(l){3-4} \cmidrule(l){5-6} \cmidrule(l){7-8} \cmidrule(l){9-10} \cmidrule(l){11-12} 
\multicolumn{2}{c}{Wilcoxon Signed-Rank Test} & \multicolumn{2}{c}{0.00128 \mytriangle{red}}  & \multicolumn{2}{c}{0.79486}              & \multicolumn{2}{c}{0.10524}     & \multicolumn{2}{c}{0.35758} & \multicolumn{2}{c}{0.33706} \\ \bottomrule
\end{tabular}
\end{table*}

\section{Results}
\label{sec:results}

In this section we will present the computational results comparing the performances of \LGAFFS{} and \PGAFFS{} in terms of the predictive accuracy and the fairness of the classification models (random forests) learned with the features selected by these GAs.

Note that comparing the performance of these two GAs in a ``fair'' way is not trivial because these two GAs are based on two different assumptions about the relative importance of different objectives: \LGAFFS{}'s lexicographic approach assumes that predictive accuracy has priority over  fairness, whilst \PGAFFS{}'s Pareto approach does not make this assumption, implicitly treating all objectives with the same priority. To further complicate the comparison between these approaches, the lexicographic GA (\LGAFFS{}) produces a single best individual once the algorithm has completed; this individual can then be used by the base classifier (random forest) and its performance measured. This is not the case for the Pareto GA (\PGAFFS{}), which produces a set of non-dominated solutions, which requires further processing or user interaction.

Hence, in order to perform a ``fair comparison'' between \LGAFFS{} and \PGAFFS{}, in the next two Subsections we report the results from two different perspectives, the lexicographic and the Pareto perspectives, as follows.

First, in Subsection \ref{sec:results_lexic_perspective}, we compare the performances of the two GAs from the lexicographic perspective. In this scenario, where predictive accuracy has priority over fairness, we reduce the Pareto set of solutions output by \PGAFFS{} to a single solution (individual) through a post-processing lexicographic filter, where that Pareto set undergoes the same lexicographic ranking that is performed for implementing elitism during each iteration of \LGAFFS{}. The best individual from the Pareto set according to this lexicographic filter is then compared to the single solution outputted by \LGAFFS{}.

Second, in Subsection \ref{sec:results_Pareto_perspective}, we compare the performances of the two GAs from the Pareto perspective, with no priority between accuracy and fairness. In this scenario, we compare the single solution output by \LGAFFS{} (here called the ``lexicographic solution'' for short) to each of the solutions in the Pareto set output by \PGAFFS{}. Then, we measure the number of solutions in the Pareto set that are dominated (in the Pareto sense) by the single lexicographic solution and, vice-versa, the number of solutions in the Pareto set that dominate the single lexicographic solution.

We emphasise that, intuitively, the lexicographic and Pareto evaluation perspectives are expected to favour the corresponding versions of the GA, and so it is important to consider both these perspectives. That is, intuitively, the lexicographic evaluation perspective is expected to lead to better results for the lexicographic GA, whilst the Pareto evaluation perspective is expected to lead to better results for the Pareto GA. Surprisingly, however, the lexicographic GA turned out to produce overall better results than the Pareto GA from both perspectives, as reported in the next two subsections.

\subsection{Comparing LPGAFFS{} and \PGAFFS{} from the Lexicographic Perspective (Prioritising Accuracy Over Fairness)}
\label{sec:results_lexic_perspective}

Table \ref{tab:comparison_lex} reports the predictive accuracy and fairness results from the lexicographic perspective. As explained earlier, to implement this perspective, the single solution returned by \LGAFFS{} is compared against the ``best'' solution output by \PGAFFS{}, which is selected by applying the lexicographic filter to the Pareto set of solutions output by that GA. Recall also that these are results for the random forest models learned with the features selected by \LGAFFS{} vs. \PGAFFS{}.

We can see in this table that the individual output by \LGAFFS{} largely outperformed the selected individual from the Pareto set produced by \PGAFFS{} with respect to the primary objective of predictive accuracy. The former recorded 18 wins against only 3 wins of the latter, and the results were statistically significant at the usual significance level of 0.05 (p-value = 0.00128). This is consistent with the fact that \LGAFFS{}'s lexicographic approach incorporated the background knowledge that maximising accuracy has priority over maximising fairness, whilst \PGAFFS{}'s Pareto approach did not allow the incorporation of this background knowledge. So, using the lexicographic filter (prioritising accuracy) to select a single solution from the Pareto set of solutions returned by \PGAFFS{}, in a post-processing phase, is clearly not as effective for maximising accuracy as prioritising accuracy during the search, which is the approach used by \LGAFFS{}.

The important question, however, is to what extent the prioritisation of accuracy led to a reduction of fairness. As shown in Table \ref{tab:comparison_lex}, 
there is no statistically significant difference between the two GAs when considering the four fairness measures; the fairness results achieved by \LGAFFS{} were even slightly better than the fairness results achieved by \PGAFFS{}. More precisely, \LGAFFS{} recorded 11 wins compared to 10 wins for \PGAFFS{}, for each of the four fairness metrics. So, despite the prioritisation of accuracy over fairness incorporated in \LGAFFS{}, it is interesting that this lexicographic approach still managed to achieve fairness results statistically equivalent to (and even slightly better regarding number of wins) the fairness results obtained by the Pareto approach used by \PGAFFS{}. This shows that a lower-priority objective can still be effectively optimised by the lexicographic approach.  

\begin{table*}[t]
\caption{Comparing Domination statistics for \LGAFFS{} and \PGAFFS{} algorithms. For each dataset and sensitive feature combination we present the number of \PGAFFS{}-produced individuals that dominate the \LGAFFS{} individual (\PGAFFS{} Domination column), the number of \PGAFFS{} individuals dominated by the individual produced by \LGAFFS{} (\LGAFFS{} Domination column), the number of \PGAFFS{} individuals that do not dominate and are not dominated by the \LGAFFS{} individual (No Domination column) and finally the proportion of individuals returned by \PGAFFS{} that have an equal or better accuracy than the individual returned by \LGAFFS{} (Pareto high accur. proport.). \label{tab:comparison_res}}
\centering
\footnotesize
\begin{tabular}{@{}llllll@{}}
\toprule
\multirow{2}{*}{Data Set} & Sensitive & \PGAFFS{}     & \LGAFFS{}     & No         & Pareto higher \\
                          & Feature & Domination    & Domination    & Domination & accur. proport.          \\ \midrule
Adult              & age                 & 0                 & 17                & 163           & 0.7111                      \\
Adult              & sex                 & 0                 & 109               & 131           & 0.0167                      \\
Adult              & race                & 0                 & 195               & 135           & 0.1364                      \\
German Credit      & age                 & 0                 & 20                & 430           & 0.1889                      \\
German Credit      & gender              & 0                 & 98                & 222           & 0.0188                      \\
Student Maths      & age                 & 0                 & 6                 & 214           & 0.0045                      \\
Student Maths      & dalc                & 1                 & 21                & 178           & 0.0550                      \\
Student Maths      & walc                & 0                 & 18                & 102           & 0.1000                      \\
Student Maths      & famrel              & 5                 & 2                 & 103           & 0.0636                      \\
Student Maths      & romantic            & 4                 & 22                & 134           & 0.2438                      \\
Student Maths      & sex                 & 16                & 0                 & 94            & 0.1818                      \\
Student Portuguese & age                 & 2                 & 101               & 207           & 0.2097                      \\
Student Portuguese & dalc                & 1                 & 22                & 297           & 0.5406                      \\
Student Portuguese & walc                & 15                & 4                 & 271           & 0.1793                      \\
Student Portuguese & famrel              & 7                 & 67                & 366           & 0.1068                      \\
Student Portuguese & romantic            & 0                 & 24                & 236           & 0.1846                      \\
Student Portuguese & sex                 & 3                 & 57                & 370           & 0.1279                      \\
Communities        & race                & 1                 & 17                & 82            & 0.6100                      \\
Default of Credit  & sex                 & 1                 & 9                 & 210           & 0.1409                      \\
Recidivisim        & race                & 0                 & 75                & 165           & 0.0667                      \\
Recidivisim        & sex                 & 0                 & 25                & 95            & 0.0000                      \\ \bottomrule
\end{tabular}
\end{table*}

\subsection{Comparing LPGAFFS{} and \PGAFFS{} from the Pareto Perspective (No Priority between Accuracy and Fairness)}
\label{sec:results_Pareto_perspective}

Table \ref{tab:comparison_res} reports the predictive accuracy and fairness results from the Pareto perspective, where there is no priority between accuracy and fairness, for each combination of a dataset and a sensitive feature. As explained earlier, in this scenario the single solution output by \LGAFFS{} (the ``lexicographic solution'') is compared against each of the solutions in the Pareto set output by \PGAFFS{}, with the comparison being based on the Pareto dominance concept. 

As shown in Table \ref{tab:comparison_res}, in 18 of the 21 classification problems (combinations of a dataset and sensitive feature), the number of Pareto solutions produced by \PGAFFS{} that are $dominated$ $by$ the lexicographic solution produced by \LGAFFS{} (column LGAFFS Domination) is greater than the number of Pareto solutions that $dominate$ the lexicographic solution (column PGAFFS Domination). In addition, the values in the column PGAFFS Domination are quite small in general, being 0 in 10 of the 21 classification problems, and the maximum value in that column is 16. By contrast, the column LGAFFS Domination has the value 0 in just one classification problem, and it has values greater than 50 in 7 classification problems. 

In summary, the lexicographic solution output by \LGAFFS{} dominates (in the Pareto sense) solutions in the Pareto set output by \PGAFFS{} much more often than vice-versa.

It should be noted that in nearly all classification problems (except the combination of dataset Adult and sensitive feature race), 
the number of solutions in the Pareto set returned by \PGAFFS{} that neither are dominated by nor dominate the lexicographic solution (column No Domination) is greater than the numbers of solutions in the two previous columns in the table. It should be expected that a large number of Pareto solutions will fall into this third category (No Domination), as the lexicographic GA concentrates mainly on a subset of the Pareto front, namely the tail containing solutions that perform very well in terms of accuracy, ignoring the other end of the Pareto front of fair solutions that potentially sacrifice accuracy.

The final column of Table \ref{tab:comparison_res} reports the proportion of individuals in the Pareto front returned by \PGAFFS{} that have an accuracy equal to or better than the accuracy of the individual returned by the lexicographic GA (\LGAFFS{}). In 18 of the 21 cases the value in that column is relatively small, meaning that the majority of the individuals in the Pareto front do not outperform the individual returned by the lexicographic GA with regard to accuracy. This should be expected because the Pareto GA spreads its resources into discovering solutions along the entire length of the front and not just one single area. Unlike the lexicographic GA, which concentrates on the area of the search space that produces solutions that are always ``good'' with respect to the highest-priority objective, which in this case is accuracy.

\section{Discussion}
\label{sec:Discussion}

We have compared two GAs for fair feature selection (in the classification task of machine learning) using two different multi-objective optimisation approaches: (a) the \LGAFFS{} algorithm \cite{brookhouse2022}, using the lexicographic approach; and (b) the \PGAFFS{} algorithm, which is a somewhat modified version of the NSGA-II proposed in \cite{Rehman2022}, using the Pareto approach. We have investigated the benefits of both approaches when selecting features for optimising both the accuracy and fairness of random forest classifiers trained with the selected features. 

Both \LGAFFS{} and \PGAFFS{} optimise the same objectives, one accuracy-based measure ($GM_{Sen \times Spec}$) and four fairness measures that capture different aspects of fairness; and they both use broadly the same individual representation, genetic operators and parameter settings (for the parameters that they have in common). So, the comparison was carefully controlled to shed light on the differences of results due to the use of the lexicographic vs. Pareto approaches.

We compared the results of these two GAs from two perspectives: a lexicographic perspective, representing the scenario where the user considers that accuracy has priority over fairness (a prioritisation incorporated into \LGAFFS{}); and a Pareto perspective, where the user does not assign any priority to the objectives of accuracy and fairness (the principle underlying \PGAFFS{}).

\subsection{Evaluation Based on the Lexicographic Perspective}

Recall that, intuitively, the lexicographic evaluation perspective was expected to lead to better results for the lexicographic GA, whilst the Pareto evaluation perspective was expected to lead to better results for the Pareto GA, since each of those GAs was designed specifically with the concepts of the corresponding evaluation in mind. However, these expectations were only partially satisfied, as follows.

In the evaluation of the results from the lexicographic perspective, we evaluated the results for each of the two objectives (accuracy and fairness) separately. In this evaluation, the lexicographic GA achieved overall significantly higher predictive accuracy than the Pareto GA. This was expected, since the former prioritises optimising accuracy over fairness. On the other hand, since the lexicographic GA assigns lower priority to fairness than to accuracy, intuitively we would expect the Pareto GA to achieve overall better fairness values. However, this was not observed, there was no statistical significance in the differences of fairness between the two GAs, and the lexicographic GA even obtained (surprisingly) slightly better fairness values than the Pareto GA, overall. 

In summary, the significant gain in accuracy associated with the lexicographic approach was achieved without sacrificing fairness.

\subsection{Evaluation Based on the Pareto Perspective}

In the evaluation of the results from the Pareto perspective, we evaluated the results in terms of optimising both objectives together, based on the concept of Pareto dominance (without prioritising any objective), by counting how often the solutions produced by one version of the GA dominates or are dominated by the solutions produced by the other version of the GA. This evaluation perspective intuitively should favour the Pareto GA, rather than the lexicographic GA. However, interestingly, the results involved two kinds of patterns, different from that initial intuition. 

First, in the large majority of cases, the solutions in the Pareto set returned by the Pareto GA neither are dominated by nor dominate the solution returned by the lexicographic GA. This first pattern suggests that, in the majority of cases, the two GAs have equivalent performance from a Pareto perspective. The second pattern, however, involves the cases where the first pattern was not observed. In these cases, the solution returned by the lexicographic GA dominated (in the Pareto sense) the solutions returned by the Pareto GA much more often than the solutions returned by the Pareto GA dominated the solutions returned by the lexicographic GA. 

In summary, in the large majority of cases, the lexicographic GA and the Pareto GA have equivalent performance in terms of Pareto dominance, but in the minority of cases where their Pareto-dominance performance differs, surprisingly, the lexicographic GA outperformed the Pareto GA.

\section{Conclusion}
\label{sec:Conclusions}

We hypothesise that the lexicographic approach was a more efficient search algorithm as its additional knowledge of the relative importance of the objectives concentrated its search in one area of the Pareto front, whereas the Pareto approach tried to find the entire Pareto front, compromising its ability to find optimal solutions in all areas.

If the user does not specify a ranking of objectives' priorities, intuitively the Pareto approach would be more natural. However, interestingly, even when evaluating the results from the Pareto perspective (with no priority between accuracy and fairness), the lexicographic GA still achieved overall better results than the Pareto GA. This suggests that the lexicographic approach deserves more attention in the multi-objective GA literature (which is currently dominated by the Pareto approach).

One open research direction includes alternative approaches to cope with many fairness objectives; since \LGAFFS{} and \PGAFFS{} currently aggregate four fairness objectives into a single fairness objective (to create a two-objective problem, optimising fairness and accuracy). In the case of \LGAFFS{}, intuitively a priority list with many separate fairness objectives could make it hard for the GA to optimise the least-priority fairness objective, and there is still the open issue of deciding the relative priorities of different fairness measures. In the case of \PGAFFS{}, care must be taken to avoid an unreasonably large number of solutions in the Pareto front as the number of fairness objectives increases.

\section*{Acknowledgements}

This work was funded by a research grant from The Leverhulme Trust, UK (Ref. No. RPG-2020-145).

\bibliographystyle{IEEEtran}
\bibliography{references.bib}

\begin{thebibliography}{10}
\providecommand{\url}[1]{#1}
\csname url@samestyle\endcsname
\providecommand{\newblock}{\relax}
\providecommand{\bibinfo}[2]{#2}
\providecommand{\BIBentrySTDinterwordspacing}{\spaceskip=0pt\relax}
\providecommand{\BIBentryALTinterwordstretchfactor}{4}
\providecommand{\BIBentryALTinterwordspacing}{\spaceskip=\fontdimen2\font plus
\BIBentryALTinterwordstretchfactor\fontdimen3\font minus
  \fontdimen4\font\relax}
\providecommand{\BIBforeignlanguage}[2]{{%
\expandafter\ifx\csname l@#1\endcsname\relax
\typeout{** WARNING: IEEEtran.bst: No hyphenation pattern has been}%
\typeout{** loaded for the language `#1'. Using the pattern for}%
\typeout{** the default language instead.}%
\else
\language=\csname l@#1\endcsname
\fi
#2}}
\providecommand{\BIBdecl}{\relax}
\BIBdecl

\bibitem{Freitas2014}
A.~Freitas, ``Comprehensible classification models: a position paper,''
  \emph{ACM SIGKDD Explorations Newsletter}, vol.~15, no.~1, pp. 1--10, 2014.

\bibitem{Burkart2021}
N.~Burkart and M.~Huber, ``A survey on the explainability of supervised machine
  learning,'' \emph{Journal of Machine Learning Research}, vol.~70, pp.
  245--317, 2021.

\bibitem{mehrabi2019survey}
N.~Mehrabi, F.~Morstatter, N.~Saxena, K.~Lerman, and A.~Galstyan, ``A survey on
  bias and fairness in machine learning,'' \emph{arXiv preprint
  arXiv:1908.09635}, 2019.

\bibitem{Binns2018}
R.~Binns, ``Fairness in machine learning: lessons from political philosophy,''
  \emph{Journal of Machine Learning Research}, vol.~81, pp. 1--11, 2018.

\bibitem{vanGiffen2022}
B.~van Giffen, D.~Herhausen, and T.~Fahse, ``Overcoming the pitfalls and perils
  of algorithms: a classification of machine learning biases and mitigation
  methods,'' \emph{Journal of Business Research}, vol. 144, pp. 93--106, 2022.

\bibitem{propublica2016}
\BIBentryALTinterwordspacing
J.~Angwin, J.~Larson, S.~Mattu, and L.~Kirchner, ``Machine bias: There’s
  software used across the country to predict future criminals. and it’s
  biased against blacks.'' 2016. [Online]. Available:
  \url{https://www.propublica.org/article/machine-bias-risk-assessments-in-criminal-sentencing}
\BIBentrySTDinterwordspacing

\bibitem{verma2018fairness}
S.~Verma and J.~Rubin, ``Fairness definitions explained,'' in \emph{2018
  IEEE/ACM International Workshop on Software Fairness (FairWare)}.\hskip 1em
  plus 0.5em minus 0.4em\relax IEEE, 2018, pp. 1--7.

\bibitem{kleinberg2016inherent}
J.~Kleinberg, S.~Mullainathan, and M.~Raghavan, ``Inherent trade-offs in the
  fair determination of risk scores,'' \emph{arXiv preprint arXiv:1609.05807},
  2016.

\bibitem{chouldechova2017fair}
A.~Chouldechova, ``Fair prediction with disparate impact: A study of bias in
  recidivism prediction instruments,'' \emph{Big data}, vol.~5, no.~2, pp.
  153--163, 2017.

\bibitem{Rehman2022}
A.~U. Rehman, A.~Nadeem, and M.~Z. Malik, ``Fair feature subset selection using
  multiobjective genetic algorithm,'' in \emph{Proceedings of the GECCO-22
  Companion (Genetic and Evolutionary Computation Conference)}.\hskip 1em plus
  0.5em minus 0.4em\relax New York: ACM Press, 2022, pp. 360--363.

\bibitem{brookhouse2022}
J.~Brookhouse and A.~Freitas, ``Fair feature selection with a lexicographic
  multi-objective genetic algorithm,'' in \emph{Proceedings of PPSN 2022:
  Parallel Problem Solving from Nature - PPSN XVII, LNCS}, vol. 13399.\hskip
  1em plus 0.5em minus 0.4em\relax Berlin: Springer International Publishing,
  2022, pp. 151--163.

\bibitem{calders2010three}
T.~Calders and S.~Verwer, ``Three naive bayes approaches for
  discrimination-free classification,'' \emph{Data Mining and Knowledge
  Discovery}, vol.~21, no.~2, pp. 277--292, 2010.

\bibitem{zemel2013learning}
R.~Zemel, Y.~Wu, K.~Swersky, T.~Pitassi, and C.~Dwork, ``Learning fair
  representations,'' in \emph{International Conference on Machine Learning},
  2013, pp. 325--333.

\bibitem{srinivas1994nsga}
N.~Srinivas and K.~Deb, ``Muiltiobjective optimization using nondominated
  sorting in genetic algorithms,'' \emph{Evolutionary computation}, vol.~2,
  no.~3, pp. 221--248, 1994.

\bibitem{deb2002nsga2}
K.~Deb, A.~Pratap, S.~Agarwal, and T.~Meyarivan, ``A fast and elitist
  multiobjective genetic algorithm: Nsga-ii,'' \emph{IEEE transactions on
  evolutionary computation}, vol.~6, no.~2, pp. 182--197, 2002.

\bibitem{valdivia2021}
A.~Valdivia, J.~Sanchez-Monedero, and J.~Casillas, ``How fair can we go in
  machine learning? assessing the boundaries of accuracy and fairness,''
  \emph{International Journal of Intelligent Systems}, vol.~36, no.~4, pp.
  1619--1643, 2021.

\bibitem{Dandl2022}
S.~Dandl, F.~Pfisterer, and B.~Bischl, ``Multi-objective counterfactual
  fairness,'' in \emph{Proc. of the GECCO'22 Companion (Genetic and
  Evolutionary Computation Conference)}.\hskip 1em plus 0.5em minus 0.4em\relax
  New York: ACM Press, 2022, pp. 328--331.

\bibitem{laCava2020}
W.~La~Cava and J.~Moore, ``Genetic programming approaches to learning fair
  classifiers,'' in \emph{Proc. Genetic and Evolutionary Computation Conference
  (GECCO-2020)}, 2020, pp. 967--975.

\bibitem{kaufman1999learning}
K.~A. Kaufman and R.~S. Michalski, ``Learning from inconsistent and noisy data:
  the aq18 approach,'' in \emph{International Symposium on Methodologies for
  Intelligent Systems}.\hskip 1em plus 0.5em minus 0.4em\relax Springer, 1999,
  pp. 411--419.

\bibitem{basgalupp2009legal}
M.~P. Basgalupp, R.~C. Barros, A.~C. de~Carvalho, A.~A. Freitas, and D.~D.
  Ruiz, ``Legal-tree: a lexicographic multi-objective genetic algorithm for
  decision tree induction,'' in \emph{Proceedings of the 2009 ACM symposium on
  Applied Computing}, 2009, pp. 1085--1090.

\bibitem{corbett2018measure}
S.~Corbett-Davies and S.~Goel, ``The measure and mismeasure of fairness: A
  critical review of fair machine learning,'' \emph{arXiv preprint
  arXiv:1808.00023}, 2018.

\bibitem{hardt2016equality}
M.~Hardt, E.~Price, and N.~Srebro, ``Equality of opportunity in supervised
  learning,'' in \emph{Advances in neural information processing systems},
  2016, pp. 3315--3323.

\bibitem{kusner2017counterfactual}
M.~J. Kusner, J.~Loftus, C.~Russell, and R.~Silva, ``Counterfactual fairness,''
  in \emph{Advances in neural information processing systems}, 2017, pp.
  4066--4076.

\bibitem{Dua2019}
\BIBentryALTinterwordspacing
D.~Dua and C.~Graff, ``{UCI} machine learning repository,'' 2017. [Online].
  Available: \url{http://archive.ics.uci.edu/ml}
\BIBentrySTDinterwordspacing

\bibitem{breiman2001random}
L.~Breiman, ``Random forests,'' \emph{Machine learning}, vol.~45, pp. 5--32,
  2001.

\bibitem{Fernandez-Delgado2014}
M.~Fernandez-Delgado, E.~Cernadas, S.~Barro, and D.~Amorin, ``Do we need
  hundreds of classifiers to solve real-world classification problems?''
  \emph{Journal of Machine Learning Research}, vol.~15, pp. 3133--3181, 2014.

\bibitem{Zhang2017}
C.~Zhang, C.~Liu, X.~Zhang, and G.~Almpanidis, ``An up-to-date comparison of
  state-of-the-art classification algorithms,'' \emph{Expert Systems with
  Applications}, no.~82, pp. 128--150, 2017.

\end{thebibliography}

\end{document}